\documentclass[10pt, conference, letter]{IEEEtran}
\IEEEoverridecommandlockouts
\usepackage{graphicx} 
\usepackage{color}
\usepackage{amsmath}
\usepackage{comment}
\usepackage{multirow}
\usepackage[table,xcdraw]{xcolor}
\usepackage{hhline}

\title{VQ4SNN: Vector Quantization for Memory-Efficient FPGA Spiking Neural Networks}
\author{
\IEEEauthorblockN{Dimitrios Sekertzis and Giorgos Dimitrakopoulos }
\IEEEauthorblockA{
Integrated Circuits Lab, Electrical and Computer Engineering, Democritus University of Thrace (DUTH), Greece}}

\begin{document}

\maketitle

\begin{abstract}
Spiking Neural Networks (SNNs) offer an energy-efficient paradigm for edge AI, making them attractive for hardware acceleration. However, deploying dense SNNs on FPGAs is constrained by limited on-chip memory for synaptic weight storage. To address this bottleneck, we propose VQ4SNN, a hardware-aware architecture that reduces memory requirements through Vector Quantization (VQ)-based weight sharing. To the best of our knowledge, this is the first application of VQ to pipelined spatial-dataflow SNN accelerators on FPGAs. VQ4SNN replaces conventional weight storage with a two-level memory organization consisting of compact pointers and a shared codebook of quantized weight vectors. The proposed design integrates FPGA-aware memory mapping with analytical VQ parameter selection, enabling efficient deployment on such accelerators while preserving inference accuracy. 
The experimental results show a reduction of 52--61\% in the total number of BRAMs compared to the state-of-the-art uncompressed FPGA SNNs without increasing overall logic utilization.
\end{abstract}

\section{Introduction}
The growing energy demands of conventional Machine Learning (ML) have renewed interest in Spiking Neural Networks (SNNs), whose sparse, event-driven computation offers a more energy-efficient paradigm for AI inference. Realizing these benefits, however, requires specialized hardware, as general-purpose processors largely negate the intrinsic efficiency of SNNs. While programmable neuromorphic multicores provide flexibility~\cite{loihi}, they incur significant architectural overhead. FPGAs offer an attractive alternative by enabling \emph{model-specific} SNN accelerators~\cite{fpga-snn} tailored at design time.

The scalability of FPGA-based SNNs is fundamentally constrained by limited on-chip memory resources~\cite{modnef}. In addition to storing neuronal state, accelerators must accommodate large numbers of synaptic weights, ideally keeping the entire network on-chip. When memory capacity is exceeded, weights must be placed in off-chip memory, significantly increasing energy consumption and reducing performance. Consequently, minimizing synaptic-weight storage is a key challenge in efficient FPGA-based SNN inference.

To address this issue, prior work has adapted compression techniques originally developed for Deep Neural Networks (DNNs), including pruning, quantization, and weight sharing~\cite{pruning}. Pruning removes less significant weights, quantization reduces numerical precision, and weight sharing replaces groups of similar weights with common representative values. Extending the latter concept, Vector Quantization (VQ) compresses multi-dimensional weight blocks using a shared codebook of representative vectors~\cite{vptq}. Although VQ has recently gained attention in DNNs~\cite{eva} and has been explored for SNNs from an algorithmic in-training perspective~\cite{spikefit}, existing studies overlook the hardware-aware design considerations required for efficient FPGA deployment.

This work presents VQ4SNN, an FPGA architecture for vector-quantized SNN inference targeting highly pipelined spatial-dataflow accelerators. VQ4SNN reduces the memory footprint of synaptic weights by replacing conventional weight storage with a compact vector-quantized representation based on shared codebooks. The proposed approach enables substantial on-chip memory savings while maintaining inference accuracy, thereby improving the scalability of FPGA-based SNNs.
The main contributions of this work are as follows:
\begin{itemize}
\item To the best of our knowledge, this is the first FPGA architecture that applies VQ to pipelined spatial-dataflow SNN accelerators, enabling substantial weight compression while preserving inference accuracy.

\item We propose a compressed weight-memory organization and 
an interleaved neuron-layer execution scheme removes the multi-port memory requirements typically associated with shared codebook accesses.

\item Comprehensive FPGA evaluation against state-of-the-art spatial-dataflow FPGA SNNs demonstrates a 52--61\% reduction in BRAM utilization, without  increasing logic utilization and preserving inference accuracy.
\end{itemize}

\section{SNN Model and FPGA Architecture Template}

SNNs process spatiotemporal data using biologically inspired neuron models. Among the models proposed in the literature, the Leaky Integrate-and-Fire (LIF) neuron is commonly adopted in FPGA accelerators due to its favorable balance between computational complexity and temporal modeling capability. More biologically detailed models, such as Izhikevich and Hodgkin--Huxley, incur higher arithmetic and area overheads because of their reliance on non-linear differential equations~\cite{fpga-snn}.

Accordingly, we use the LIF model as the baseline neuron implementation throughout this work. Since VQ4SNN targets synaptic-weight storage and access, the proposed compression framework is largely orthogonal to the underlying neuron model and can be integrated into alternative SNN architectures with minimal modification.

\begin{figure}[t]
\centering
\includegraphics[width=0.58\columnwidth]{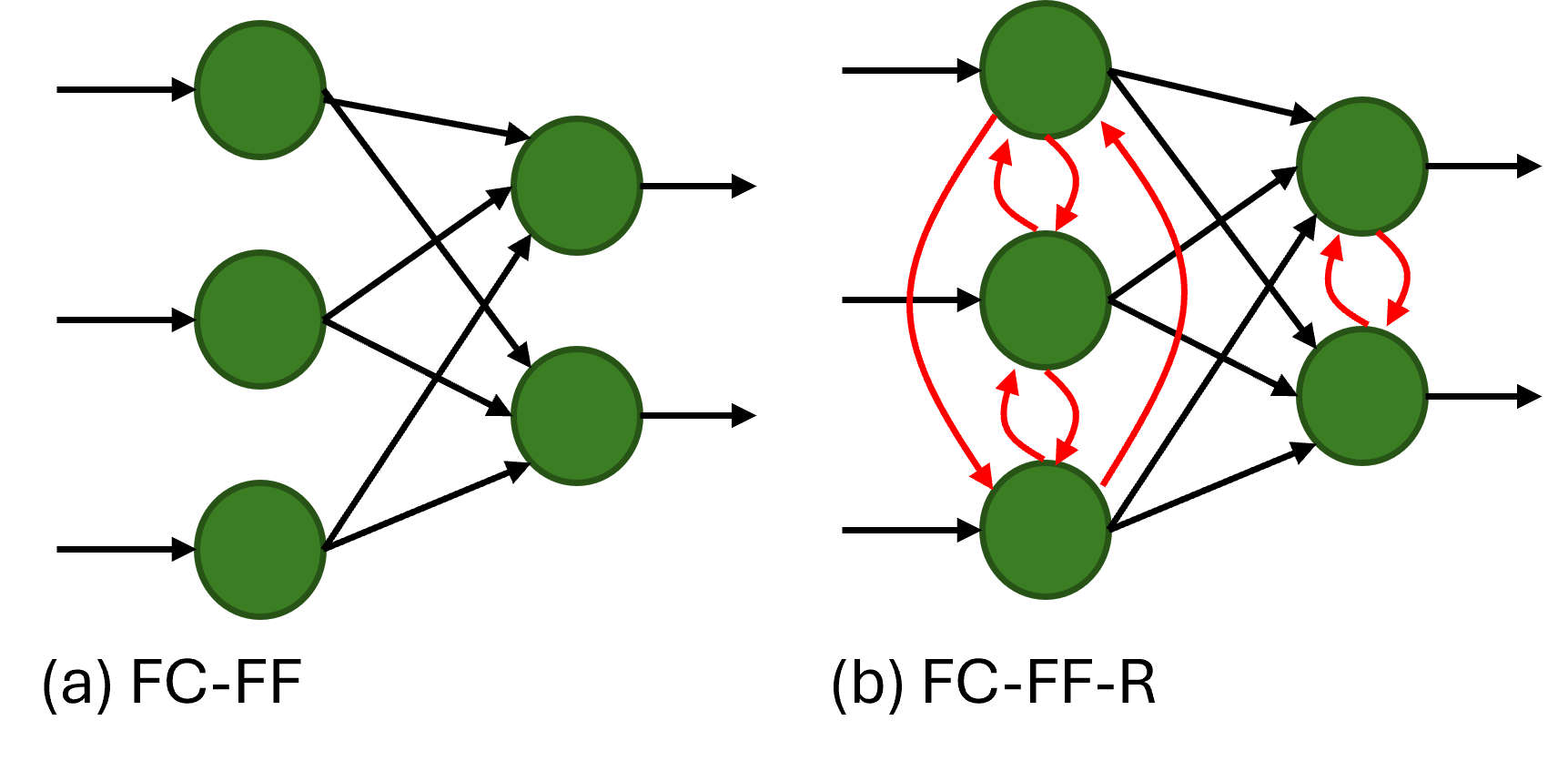}
\caption{Fully Connected (FC) SNNs with (a) feedforward (FF) connections and (b) with feedforward and intra-layer recurrent (R) synapses (in red).}
\label{f:snn}
\end{figure}

\subsection{LIF Neuron Model}
To implement continuous-time LIF dynamics on FPGAs, the governing differential equations are discretized into fixed time steps. A time step represents the minimum unit of time during which a neuron evaluates its inputs, updates its membrane potential, and potentially emits an output spike. The resulting neuron state update is given by
\begin{equation}
V(t) = V(t-1) + \sum W_i S_i(t) - \alpha V(t-1) ,
\end{equation}
where $V(t)$ denotes the membrane potential at time step $t$, $W_i$ are the synaptic weights, and $S_i(t)$ are incoming binary spikes. The leakage factor $\alpha$, derived from the membrane time constant, controls the exponential decay of the membrane potential between consecutive time steps.

A neuron emits an output spike whenever its membrane potential exceeds a threshold $V_{th}$. Following a spike, the membrane potential is reset either through a hard reset, which restores the resting potential, or a soft reset, which subtracts the threshold while preserving residual charge.

Fig.~\ref{f:snn} illustrates fully connected SNN topologies. The most common organization consists of layers of neurons connected through feedforward synapses (Fig.~\ref{f:snn}(a)). Some architectures further incorporate intra-layer inhibitory connections (Fig.~\ref{f:snn}(b)), whereby neurons suppress the activity of neighboring neurons through negative synaptic weights.

\subsection{Input Encoding}
Static numerical inputs are commonly converted into spike trains using firing-rate encoding~\cite{fpga-snn}. Unlike temporal encoding schemes that rely on precise spike timing, rate encoding represents an input value as the probability of generating a spike at each time step. Consequently, larger input values produce denser spike trains over the observation window. This simple and hardware-efficient encoding mechanism integrates naturally with discrete-time SNN inference.

\subsection{Spatial-Dataflow FPGA SNN Accelerator}

\begin{figure}
\centering
\includegraphics[width=0.88\columnwidth]{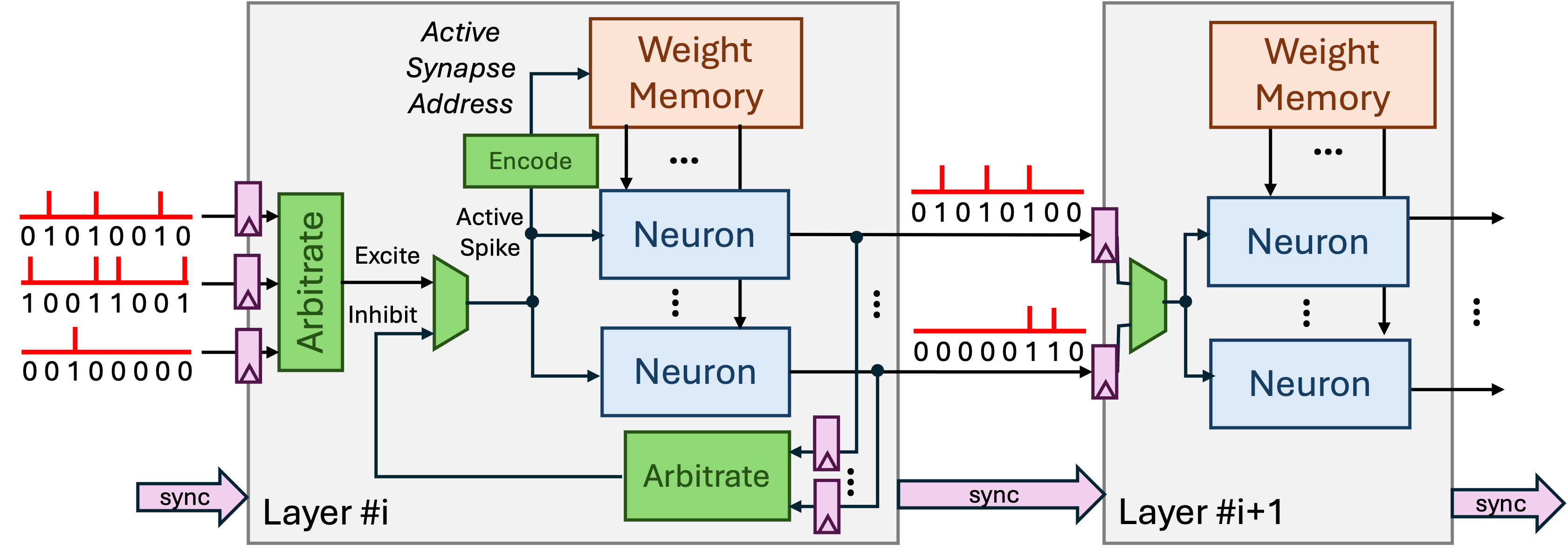}
\caption{The organization of Spiker spatial dataflow FPGA SNN accelerator with arbitrated active spike selection.}
\label{f:spiker}
\end{figure}

Numerous FPGA architectures have been proposed for SNN inference~\cite{fpga-snn,modnef,spiker_plus,firefly}. However, this work focuses on spatial-dataflow accelerators, in which the entire network topology is instantiated directly in hardware. A representative and resource-efficient example is the Spiker architecture~\cite{spiker_plus} shown in Fig.~\ref{f:spiker}, which serves as the baseline platform for the proposed VQ4SNN framework.

Spiker is characterized by clock-driven state update and a sequential spike-processing model. Within each time step, input spikes are arbitrated and processed individually. Whereas the original design employed static spike multiplexing, our implementation uses spike arbitration as an equivalent lower-latency alternative. For each selected spike, the corresponding synaptic-weight row is fetched from weight memory and broadcast to all neurons in the target layer, enabling parallel membrane-potential updates within a single clock cycle.

Neurons first process excitatory spikes received from the preceding layer and subsequently process intra-layer inhibitory spikes using the same mechanism. This execution model ensures that each neuron performs exactly one arithmetic operation per active cycle: either a weight addition for excitation or a weight subtraction for inhibition.

After processing all spikes of the current time step, the layer enters a state-evaluation phase. Membrane potentials are compared with the threshold to generate output spikes. Neurons that fire are reset to a predefined value. Then membrane leakage is applied using binary right shifts. In~\cite{spiker_plus} it was shown that approximating decay factors by negative powers of two has negligible impact on inference accuracy. Execution advances to the next time step only after all layers have completed processing of the current one.

\begin{figure}[t]
\centering
\includegraphics[width=0.78\columnwidth]{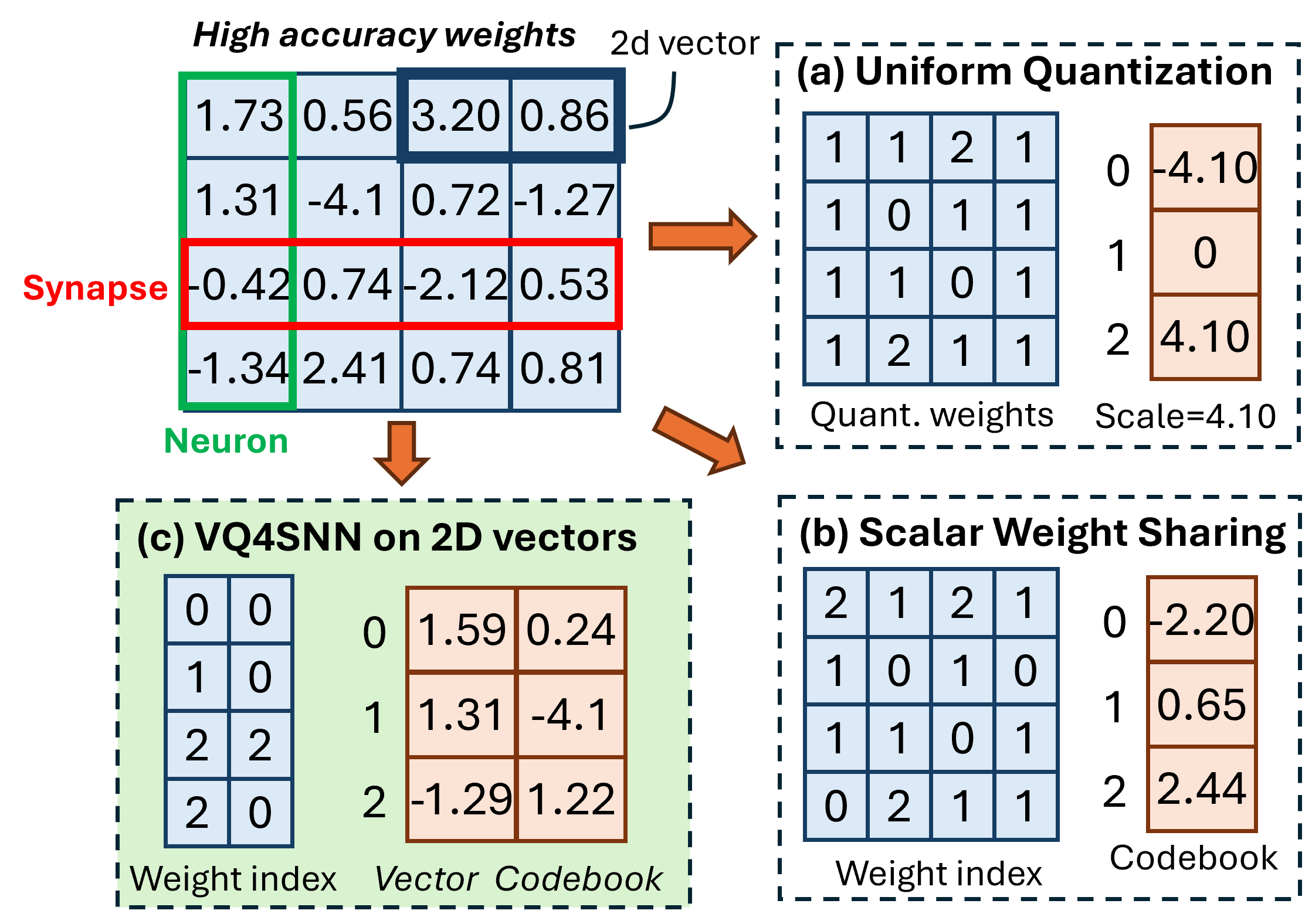}
\caption{Weight compression: (a) Quantization reduces weight precision. (b) Scalar weight sharing replaces individual weights with codebook indices. (c) VQ4SNN applies VQ to pairs of consecutive synaptic weights.}
\label{f:quant}
\end{figure}

\section{Weight compression with vector quantization}
Neural-network weight compression is commonly achieved through quantization~\cite{flowq} and weight sharing, which are often applied jointly to maximize memory efficiency. Quantization reduces storage by mapping weights to low-bit numerical representations (e.g., INT4 or INT8) using predefined arithmetic transformations, as illustrated in Fig.~\ref{f:quant}(a). Weight sharing further compresses the model by representing similar weights with a common value selected from a shared codebook and replacing individual weights with compact codebook indices (Fig.~\ref{f:quant}(b)). In practice, the codebook is typically generated post-training using clustering algorithms such as K-means.

\subsection{Vector Quantization in VQ4SNN}
VQ4SNN builds on these principles by employing VQ as its primary compression mechanism. Instead of sharing individual weights, VQ4SNN groups consecutive synaptic weights into $d$-dimensional vectors and replaces each vector with a compact codebook index. The codebook entries themselves remain quantized and are stored using the target weight precision, making VQ complementary to conventional low-bit quantization. 
By sharing entire weight vectors rather than individual weights, VQ4SNN exploits correlations among neighboring synapses and achieves higher compression ratios than scalar weight sharing. As shown in Fig.~\ref{f:quant}(c), VQ is applied to consecutive synapses within each weight row, such that each codebook entry corresponds to a $d$-dimensional weight vector. The codebook is generated offline and post-training using K-means clustering on the original weight vectors.

A key design choice in VQ4SNN is the application of VQ on a per-layer basis rather than across the entire network. The effectiveness of vector quantization varies considerably between layers, and small layers often contain insufficient redundancy to justify the reconstruction error introduced by compression. Consequently, VQ4SNN selectively applies VQ only to layers that provide meaningful memory savings while preserving inference accuracy.

\begin{figure}[t]
\centering
\includegraphics[width=0.68\columnwidth]{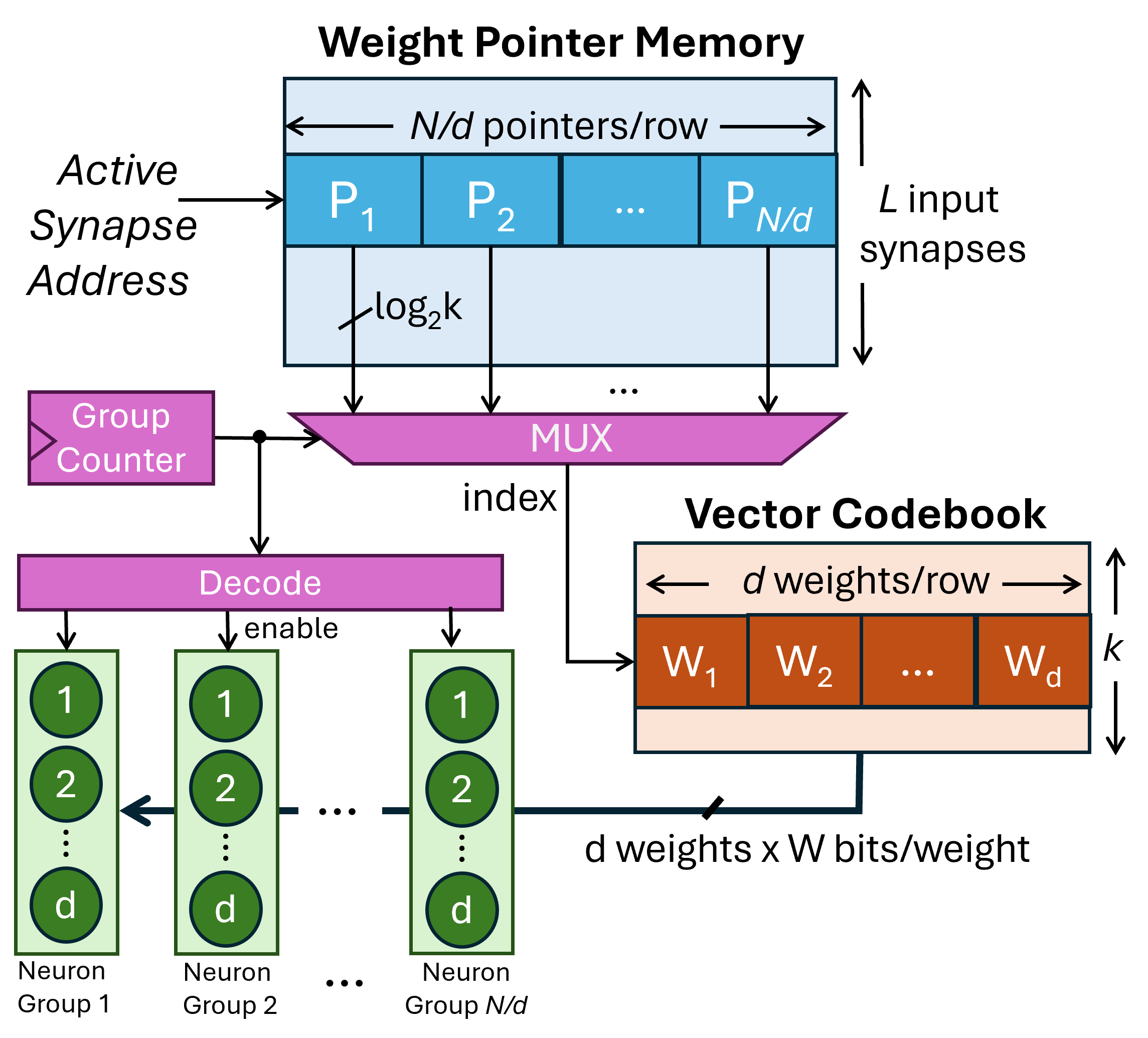}
\caption{Two-level memory organization: Original weights are replaced by pointers to vector-quantized codebook entries. Weight vectors are retrieved and applied to groups of $d$ neurons in successive phases.}
\label{f:mem}
\end{figure}

\subsection{Two-level memory microarchitecture}
VQ4SNN targets the spatial-dataflow SNN architecture of Fig.~\ref{f:spiker}, where each incoming spike triggers a wide memory access that retrieves the synaptic weights for all $N$ neurons in the target layer, enabling parallel membrane-potential updates within a single cycle. Neuron state, including membrane potentials and firing thresholds, is stored locally in flip-flops, whereas synaptic weights reside in BRAM-based weight memories. The storage and parallel access requirements of these weights incur a substantial memory footprint. To address this challenge, VQ4SNN replaces the conventional BRAM-based weight memory with a two-level hierarchy comprising a compact pointer memory and a shared vector-quantization codebook, depicted in Fig.~\ref{f:mem}.

Applying VQ to blocks of $d$ consecutive weights compresses each synaptic row from $N$ weights to $N/d$ pointers. However, preserving the original fully parallel update mechanism would require the shared codebook to service $N/d$ accesses simultaneously, effectively requiring a highly multi-ported memory. For dense SNN layers, such a design is impractical on FPGAs due to its area and routing overhead.

To address this challenge, VQ4SNN adopts an interleaved execution strategy. Rather than updating all $N$ neurons simultaneously, neurons are partitioned into groups of $d$. Upon receiving an input spike, a counter iterates through the $N/d$ pointers associated with the corresponding synaptic row. Each pointer accesses the shared codebook and retrieves a $d$-dimensional weight vector, which is accumulated into the membrane potentials of the corresponding neuron group. Consequently, the update of all $N$ neurons is distributed across $N/d$ cycles, or $N/(2d)$ cycles when the codebook can exploit the native dual-port capability of FPGA BRAMs. All neurons remain physically instantiated in hardware, but only the currently addressed group is active, while the remaining neurons are clock-gated. This organization preserves the deterministic dataflow execution model of the original architecture while replacing the prohibitive requirement for $N/d$ simultaneous codebook accesses with time-interleaved accesses. The resulting latency overhead depends on spike activity density and is evaluated experimentally in Section~\ref{s:eval}.

\section{Evaluation}
\label{s:eval}
\begin{table*}[t]
\centering
\caption{Performance and resource comparison of VQ4SNN, Spiker+~\cite{spiker_plus}, and ModNef~\cite{modnef} FPGA SNNs.}
\setlength{\tabcolsep}{4pt}
\renewcommand{\arraystretch}{1.1}
\begin{tabular}{|c||c|c|c||c|c||c|c||c|c|}\hline
Dataset
& \multicolumn{3}{c||}{\bf MNIST Base} & \multicolumn{2}{c||}{\bf SHD} & \multicolumn{2}{c||}{\bf AudioMNIST} & \multicolumn{2}{c|}{\bf MNIST Large} \\ 
(SNN Structure)& \multicolumn{3}{c||}{(784-128-10)}  & \multicolumn{2}{c||}{(700-200-20)} & \multicolumn{2}{c||}{(40-150-10)} & \multicolumn{2}{c|}{(784-392-196-10)} \\ \hline
\multirow{3}{*}{Design}
& \multirow{3}{*}{Spiker+} & \multirow{3}{*}{ModNEF} & Proposed & \multirow{3}{*}{Spiker+}& Proposed & 
\multirow{3}{*}{Spiker+} & Proposed & 
\multirow{3}{*}{ModNEF}  & Proposed \\ 
& & & L1:$d$=8 $k$=$2^{11}$ & 
 & L1:$d$=4 $k$=$2^{11}$ & 
 & L1:$d$=4 $k$=$2^{10}$ & 
 & L1:$d$=8 $k$=$2^{11}$ \\ 
& & & & & & & & & L2:$d$=8 $k$=$2^{11}$ \\ \hline\hline
Network Type & FC-FF & FC-FF & FC-FF-R & FC-FF-R & FC-FF-R & FC-FF & FC-FF-R & FC-FF & FC-FF-R \\ \hline
Ref. Accuracy & 96.83\% & 95.51\% & 96.35\% & 75.44\% & 73.00\% & 95.23\% & 93.12\% & 93.20\% & 94.38\%\\ \hline
HW Accuracy & 93.85\% & 95.76\%  & 95.29\% & 72.99\% & 72.00\% & 89.82\% & 90.60\% & 96.96\% & 92.70\% \\ \hline
Weight bits & 4 & 10 & 5 & F(6)/R(5) & 7 & 5 & 6 & 8 & 8 \\ \hline
Memb. Potential bits & 6 & N.R.$^a$ & 11 & 8 & 14 & 8 & 12 & 16 & 15 \\ \hhline{==========}
LUTs & 4314 & 5947 & 7055 &  & 13628 & & 7562 & 20321 & 33081 \\ \hhline{----~-~---}
FFs & 3298 & 3060 & 3606 & \multirow{-2}{*}{18268} & 5686 & \multirow{-2}{*}{10124} & 3375 & 28284 & 14809 \\ \hline
DSPs & - & - & - & - & - & - & - & 3 & - \\ \hline
BRAMs & 18.0 & 42.5  & 8.5 & 51.0 & 19.5 & 16.0 & 7.5 & 113.5 & 47.5 \\ \hline
Power (mW) & 180 & 141 & 158 & 430 & 250 & 290 & 185 & 336 & 364 \\ \hhline{==========}
Latency/input$^b$ (ms) & 0.780 & 0.115 & 0.246 & 0.540 & 2.412 & 0.080 & 0.632 & 0.904 & 1.093 \\ \hline
Spikes time-steps & 100 & 100 & 25 & 100 & 100 & 73 & 100 & 100 & 25 \\ \hline
\end{tabular}
$^a$N.R.: Not Reported \qquad $^b$Input refers to one complete inference sample (e.g., a single image or audio recording)
\label{t:eval}
\vspace{-12pt}
\end{table*}

To evaluate VQ4SNN, we compare it against Spiker+~\cite{spiker_plus} on MNIST~\cite{mnist}, SHD~\cite{shd}, and Audio MNIST~\cite{audiomnist}, and against ModNef~\cite{modnef} on two MNIST configurations of different sizes. Spiker+ and ModNef were selected because they represent highly resource-efficient spatial-dataflow FPGA SNN accelerators, making them strong baselines for assessing the effectiveness of the proposed approach. All designs target a Xilinx Artix-7 (xc7v020) FPGA operating at 100 MHz.

For VQ4SNN, all models were trained using SNNTorch~\cite{snntorch} and implemented in SystemVerilog RTL. Input encoding follows~\cite{spiker_plus} for MNIST and Audio MNIST, and~\cite{shd} for SHD. The complete training flow, RTL implementation, and experimental framework are open-sourced at {\tt\footnotesize https://github.com/ic-lab-duth}, while results for Spiker+ and ModNef are taken directly from their published works.

Table~\ref{t:eval} summarizes the network configurations and implementation results. The same network sizes, reported in the first row, are used for each dataset with the only architectural variation being the presence of recurrent inhibitory connections following network types of Fig.~\ref{f:snn}.
The table reports both the full-precision reference accuracy and the accuracy achieved in hardware by the implemented FPGA SNNs. For VQ4SNN, the selected compression settings, including the compressed layers and corresponding $(d,k)$ parameters, are also reported.

Across all benchmarks, VQ4SNN preserves inference accuracy while maintaining LUT, FF, and power consumption comparable to or lower than the baseline architectures. For Spiker+, only the combined LUT+FF utilization is available for SHD and Audio MNIST.

The primary benefit is the substantial reduction in BRAM usage, reaching 52\%, 61\%, and 53\% for MNIST, SHD, and Audio MNIST, respectively, when compared to Spiker+. 
These savings can be explained using MNIST as an example, where VQ is applied to 
the 5-bit weights of 
Layer~1 with $(d,k)\!=\!(8,2048)$. This layer contains 912 synaptic rows (784 feedforward and 128 inhibitory), each compressed into $N/d=16$ pointers of $\log_2 2048\!=\!11$ bits. 
Assuming standard 1024$\times$36-bit BRAMs, the 912 pointer rows ($16 \times 11 = 176$ bits/row) map to 5 BRAMs. The shared codebook (2048 vectors of $8 \times 5 = 40$ bits/vector) requires 2.5 BRAMs to support dual-porting (two 36Kbit and one 18Kb BRAM), with 1 additional BRAM for the uncompressed output layer. Overall, VQ4SNN consumes just 8.5 BRAMs, while Spiker+ with smaller 4-bit weights spends 18 BRAMs for an SNN with the same size and without any inhibitory weights.

VQ4SNN exhibits a latency tradeoff dependent on spike activity density.
VQ4SNN achieves lower latency for MNIST, but exhibits higher latency for SHD and Audio MNIST. The increase originates from the time-shared use of the vector codebook, which distributes the evaluation of a layer across multiple cycles. Although this overhead is partially mitigated by arbitrating active spikes, it becomes noticeable for workloads with high spike activity. Importantly, this latency increase is not an inherent limitation of the proposed approach. The substantial BRAM savings achieved by VQ4SNN leave ample room for codebook replication, enabling additional parallel accesses that reduce latency while still retaining most of the memory savings.

The final two columns of Table~\ref{t:eval} compare VQ4SNN against ModNef on a larger MNIST network. By applying VQ independently to two layers with different $(d,k)$ configurations, VQ4SNN reduces BRAM usage by 58\% while maintaining model accuracy and logic utilization comparable to ModNef under the same weight bitwidth.

\section{Conclusions}
In this paper, we presented VQ4SNN, an FPGA SNN architecture that addresses the memory bottlenecks of SNN inference through weight VQ. By combining a compressed two-level memory organization with a time-shared codebook access scheme VQ4SNN significantly reduces on-chip storage requirements with the need for heavy multiporting. Hardware evaluations on representative SNN workloads demonstrate substantial reductions in overall memory footprint and BRAM utilization while preserving baseline inference accuracy. 

As future work, we plan to co-optimize SNN training and compression by encouraging similarity among neighboring synaptic weights, and also to improve neuron microarchitecture to further optimize FPGA resource utilization.

\bibliographystyle{IEEEtran}
\bibliography{refs}

\end{document}